\documentclass[10pt, a4paper, twocolumn]{IEEEtran}

\usepackage[utf8]{inputenc}
\usepackage{hyperref}
\usepackage{graphicx}
\usepackage{float}
\usepackage{booktabs}
\usepackage{wrapfig}
\usepackage{subcaption}
\usepackage{caption}
\usepackage[section]{placeins}
\usepackage{eurosym}
\usepackage{xcolor}
\usepackage{footnote}
\usepackage{authblk}

\begin{document}
\title{\textbf{Dataset shift quantification for credit
card fraud detection}}



\author[1,2]{Yvan Lucas}
\author[1]{Pierre-Edouard Portier}
\author[1]{Léa Laporte}
\author[1]{Sylvie Calabretto}
\author[3]{Liyun He-Guelton}
\author[3]{Frederic Oble}
\author[2]{Michael Granitzer}

\affil[1]{LIRIS UMR5205}
\affil[2]{Universitat Passau}
\affil[3]{Worldline Lyon}

\renewcommand\Authands{ and }
%

\maketitle

\begin{abstract}
Machine learning and data mining techniques have been used extensively in order to detect credit card frauds. However purchase behaviour and fraudster strategies may change over time. This phenomenon is named dataset shift \cite{torres2012} or concept drift in the domain of fraud detection \cite{abdallah2016}.

In this paper, we present a method to quantify day-by-day the dataset shift in our face-to-face credit card transactions dataset  (card holder located in the shop) . In practice, we classify the days against each other and measure the efficiency of the classification. The more efficient the classification, the more different the buying behaviour between two days, and vice versa. Therefore, we obtain a distance matrix characterizing the dataset shift.

After an agglomerative clustering of the distance matrix, we observe that the dataset shift pattern matches the calendar events for this time period (holidays, week-ends, etc).

We then incorporate this dataset shift knowledge in the credit card fraud detection task as a new feature. This leads to a small improvement of the detection.

\end{abstract}


\section{Introduction}
Credit card fraud detection presents several difficulties. One of them is the fact that purchase behaviours may evolve over time. This dataset shift or concept drift may cause issues since the fraud detection system may become inadapted: the decision function it learnt on the training set does not correspond to the distribution of the testing set.

Concept drift and dataset shift have already been described in the litterature. Some works focused on characterizing the concept drift whereas others aimed to adapt to it. 

Abdallah \& al. \cite{abdallah2016} described concept drift as the phenomenon where the underlying model (or concept) is changing over time: the purchasing behaviour may evolve over time and the fraudster may adapt their techniques. In this case, the decision function learned on the training set may become out of date and not reflect the conditional probability of the target variable ($y$) given the attributes ($X$) of the testing set transactions. They distinguish two approaches for solving this issue: the evolving approach refers to learners that stay up to date with the data stream whereas the regulated approach aims to detect concept drift and take action in order to adapt the fraud detection system when there is one (usually re training the learner).

Gao \& al. \cite{gao2007} extended the notion of dataset shift to the joint distribution $P(X,y)$\footnote{In this paper, $y$ refers to the target variable: 1 if the transaction is a fraud, 0 if the transaction is genuine and $X$ refers to the set of features describing the transaction: amount, age of the card holder, merchant category, etc.} instead of only the conditional distribution $P(y|X)$: a drift in the conditional distribution $P(y|X)$ would lead to a drift in the joint distribution $P(X,y)$ since $P(X,y) = P(y|X)*P(X)$. They argue in favor of learners that may adapt over time to a change of the probablity distribution. 

Pozzolo \& al. \cite{pozzolo2015} \cite{pozzolo2014} introduced the notion of seasonality in concept drift. The purchase behavior may evolve with seasons. They proposed to use investigator's feedbacks in a sliding window approach in order to adapt the classifiers over time.  

In 2012, Torres \& al \cite{torres2012} proposed a denomination for the different variations of the distributions from the train set to the test set. The variation of the distribution $P(X)$ is named \textit{covariate shift} whereas the variation of the conditional distribution $P(y|X)$ is named \textit{concept shift}. Dataset shift embeds covariate shift and concept drift since it describe the change in the joint distribution $P(X,y)$. Their paper is independent from the context of credit card fraud detection and therefore they didn't restrain their description of dataset shift to a change over time of the distribution.	


In the credit card fraud detection litterature, the concept drift is mostly diagnosed passively when there is an increase in the error rate of the fraud detection system. In this work we present a strategy to early detect concept drift, more precisely dataset shift in a belgian credit card transactions dataset spanning the months from March to May 2015. 

We will first describe how to quantify covariate shift in section \ref{covshiftquant}, then we will cluster the covariate shift matrix in section \ref{cluster} and use these clusters for credit card fraud detection in section \ref{predict}.

\section{Covariate shift quantification}
\label{covshiftquant}


We work on a face-to-face credit card transactions dataset provided by our industrial partner. This dataset contains the anonymized transactions from all the belgian credit cards between 01.03.2015 and 31.05.2015.

The task is to predict the class of the transactions (genuine or fraudulent).

Transactions are represented by vectors of continuous, categorical and binary features that characterize the card-holder, the transaction and the terminal. In this study, the card-holder is characterized by his age, gender and bank. The transaction is characterized by its amount, type of payment (with PIN check or contactless) and other confidential features. The terminal is characterized by its country and a merchant category code.

We aim to build a distance matrix in order to quantify the possible covariate shift happening over the 92 days of the dataset.  

For this purpose we build and evaluate learners that try to tell from which day of the considered pair of days each transaction belongs. If the learner achieves to detect the day of the testing set transactions, it means that the transactions from each of the two days considered are easy to differenciate. We assume that the two days can be modeled with two distant distributions: there is a covariate shift between these two days. On the other hand, if the two days are hard to differenciate, that means that their transactions are similar: there is no covariate shift between these days.

For each pair of days, a Random Forest classifier is trained using 20000 transactions from each day (40000 in total) in order to detect the day of each transaction. Afterwards, this Random Forest classifier is asked to classify a test set containing 5000 transactions from each day (10000 in total). 

The classification is evaluated using the Matthews correlation coefficient (MCC). The Matthews correlation coefficient\footnote{$MCC = \frac{TP \times TN - FP \times FN}{\sqrt{(TP+FP)(TP+FN)(TN+FP)(TN+FN)}}$}
 is a confusion matrix based metric that can take values between -1 and 1. A value of MCC close to 1 means that the classifier is good at differenciating the pair of days considered, the days are easy to differenciate since there is a dataset shift between them.
 
In the end, this MCC value is used in order to build a 92x92 distance matrix characterizing the covariate shift between the 92 days of the dataset (see figure \ref{dist_mat} in appendix).

\section{Agglomerative clustering using the distance matrix between days}
\label{cluster}
Regularities can be observed in the covariate shift (see figure \ref{dist_mat} in appendix). We can identify 4 clusters of days that are similar to each other and dissimilar to other days. We can correlate these clusters with calendar event (Dataset of belgian transactions): 
\begin{itemize}
\item \textbf{Working days:} During these days, people are working and merchants are mostly open.
\item \textbf{Saturdays:} During these days, people are not working and merchants are mostly open.
\item \textbf{Sundays:} During these days, people are not working and merchant are mostly closed. We observe that some religious holidays (catholic) seem to share the same dataset shift (1st may: labor Day, 14th may: Ascent and 25th may: Pentecost).
\item \textbf{School holidays:} During these days, people are working and the merchants are mostly open but the children are not at school. Some parents take holidays to look after their children or travel with them. 
\end{itemize}

This clustering is based on a qualitative observation of the distance matrix between days (figure \ref{dist_mat} in appendix) and may be biased. Therefore, we do a hierarchical clustering on the distance matrix.

\begin{figure}[H]
\centering
\includegraphics[width=0.4\textwidth]{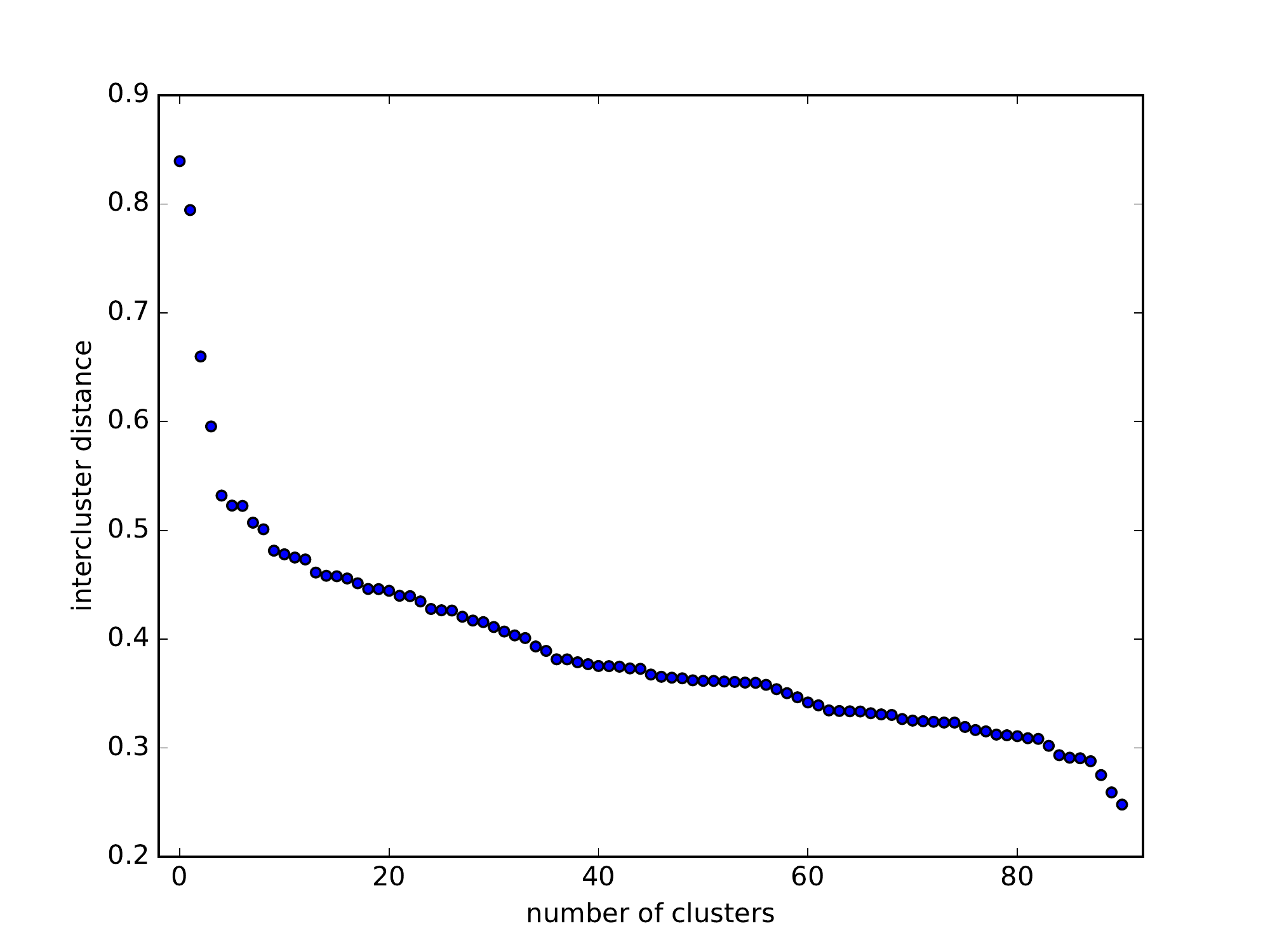}
\caption{Diminution of the average inter-cluster distance with the increase of the number of clusters}
\label{ncluster}
\end{figure}

The dendrogram created with agglomerative clustering can be cut at different levels. These levels correspond to different number of clusters. In figure \ref{ncluster} we observe that the drops sharply of the average intercluster distance decreases after 4 clusters. This information along with the previous qualitative clustering reinforces the choice of 4 clusters.

The output of the hierarchical clustering is shown in table \ref{clusters}. The 4 clusters match almost perfectly the calendar-based clusters previously identified.
\begin{description}
\item[\textcolor{red}{0}] Week days (working days)
\item[\textcolor{blue}{1}] Sundays and work holidays (Ascent, monday Pentecost and Labor day): Shops are closed
\item[\textcolor{green}{2}] Saturdays and most of the Fridays : Days of the end of the week where shops are open
\item[\textcolor{yellow}{3}] Week days of eastern school holidays
\end{description}

\begin{savenotes}
\begin{table}[h]
\begin{tabular}{l|ccccccc}
Week & MON & TUE & WED & THU & FRI & SAT & SUN\\
\hline
9 & & & & & & & 1	 \\
10 &\textcolor{red}{0}  &\textcolor{red}{0} &\textcolor{red}{0} &\textcolor{red}{0} &\textcolor{red}{0} &\textcolor{green}{2} &  \textcolor{blue}{1} \\
11 &\textcolor{red}{0} &\textcolor{red}{0} &\textcolor{red}{0} &\textcolor{red}{0} &\textcolor{green}{2} &\textcolor{green}{2} & \textcolor{blue}{1} \\
12 &\textcolor{red}{0} &\textcolor{red}{0} &\textcolor{red}{0} &\textcolor{red}{0} &\textcolor{green}{2} &\textcolor{green}{2} & \textcolor{blue}{1} \\
13 &\textcolor{red}{0} &\textcolor{red}{0} &\textcolor{red}{0} &\textcolor{red}{0} &\textcolor{green}{2} &\textcolor{green}{2} & \textcolor{blue}{1} \\
14 &\textcolor{red}{0} &\textcolor{red}{0} &\textcolor{red}{0} &\textcolor{red}{0} &\textcolor{green}{2} &\textcolor{green}{2} & \textcolor{blue}{1} \\
15 &\textcolor{blue}{1} &\textcolor{yellow}{3} &\textcolor{yellow}{3} &\textcolor{yellow}{3} &\textcolor{yellow}{3} &\textcolor{green}{2} & \textcolor{blue}{1} \\
16 &\textcolor{yellow}{3} &\textcolor{yellow}{3} &\textcolor{yellow}{3} &\textcolor{yellow}{3} &\textcolor{green}{2} &\textcolor{green}{2} & \textcolor{blue}{1} \\
17 &\textcolor{red}{0} &\textcolor{red}{0} &\textcolor{red}{0} &\textcolor{red}{0} &\textcolor{red}{0} &\textcolor{green}{2} & \textcolor{blue}{1} \\
18 &\textcolor{red}{0} &\textcolor{red}{0} &\textcolor{red}{0} &\textcolor{green}{2} &\textcolor{blue}{1 \footnote{1st may: Labor day}}&\textcolor{green}{2} & \textcolor{blue}{1} \\
19 &\textcolor{red}{0} &\textcolor{red}{0} &\textcolor{red}{0} &\textcolor{red}{0} &\textcolor{green}{2} &\textcolor{green}{2} & \textcolor{blue}{1} \\
20 &\textcolor{red}{0} &\textcolor{red}{0} &\textcolor{green}{2} &\textcolor{blue}{1\footnote{14th may: Ascent}} &\textcolor{green}{2} &\textcolor{green}{2} & \textcolor{blue}{1} \\
21 &\textcolor{red}{0} &\textcolor{red}{0} &\textcolor{red}{0} &\textcolor{red}{0} &\textcolor{green}{2} &\textcolor{green}{2} & \textcolor{blue}{1} \\
22 &\textcolor{blue}{1\footnote{25th may: Pentecost}} &\textcolor{red}{0} &\textcolor{red}{0} &\textcolor{red}{0} &\textcolor{green}{2} &\textcolor{green}{2} & \textcolor{blue}{1} \\
\end{tabular}
\caption{Agglomerative clustering of the days using the distance matrix obtained by classifying each day against every other day}
\label{clusters}
\end{table}
\end{savenotes}

The framework that consists in building a distance matrix between days then clustering using this distance matrix let us identify 4 different types of days in the dataset with a strong covariate shift between them. This covariate shift pattern matches the events of the belgian calendar for this time period.
\section{Incorporation of dataset shift knowledge}
\label{predict}
In order to leverage the knowledge of the covariate shift between days, we incorporate it as a new categorical feature with one modality for each identified cluster.

We use random forest classifier in order to evaluate our incorporation of covariate shift knowledge as an additional feature since it is the most used classifier for credit card fraud detection and adapts well to heterogeneous datasets (datasets with features of different types) \cite{bolton2001}. Random forest is an ensemble based classifier that creates C4.5 trees with a randomly chosen subset of features for each tree. The decisions of all the trees of the classifier are then aggregated by majority voting. 

We repeat 5 times an experimental protocol where the face-to-face transactions from 7 consecutive days are used as a train set  and the face-to-face transactions from 7 consecutive days with a gap of 7 days after the train set are used as a test set. We evaluate the addition of the dataset shift feature by computing the Precision-Recall AUC and the ROC AUC for the classifiers with and without the dataset shift feature.

We chose to separate the time period corresponding to the training and validation set and the time period corresponding to the testing set with a gap of 7 days. The reason is that in the real world fraud detection systems, human investigators have to manually verify the alerts generated by the classifiers. Since this process takes time, the ground truth is delayed by about one week.


\begin{table}[h]
\centering
\begin{tabular}{l|cc|cc}
 & \multicolumn{2}{c}{PR AUC} & \multicolumn{2}{|c}{ROC AUC}\\
 test set & without & with & without & with \\
\hline
 16/03-22/03 & 0.190 & 0.175 & 0.964 & 0.961 \\
 23/03-29/03 & 0.288 & 0.303 & 0.969 & 0.973 \\
 30/03-05/04 & 0.131 & 0.142 & 0.939 & 0.945 \\
 06/04-12/04 & 0.256 & 0.273 & 0.914 & 0.927 \\
 13/04-19/04 & 0.131 & 0.126 & 0.946 & 0.940 \\
 \hline
 average & 0.199 & 0.204 & 0.946 & 0.949 \\
\end{tabular}
\caption{AUC variations with the addition of the covariate shift feature for different testing periods}
\end{table} 

The random forest hyperparameters are described in table \ref{gridRF}. The same hyperparameters are used for the construction of the distance matrix between days. 

\begin{table}[h] 
\centerline{
\begin{tabular}{ccc}
n-trees & n-features & min-samples-leaf\\
\hline
$100$ & $sqrt$ & $10$ \\ 
\end{tabular}}
\caption{Random Forest hyperparameters
\label{gridRF}}
\end{table}

Adding the covariate shift information for the classification increases the precision-recall AUC by 2.5\%. The ROC AUC is also slightly increased.

However this improvement may partly be a consequence of the weak feature set used and could disappear when using stronger feature engineering strategies such as feature aggregation \cite{bahnsen2016}\cite{whitrow2008} or sequence modelling with Hidden Markov Models \cite{lucas2019} or recurrent neural networks \cite{jojo2018}.
\section*{Conclusion}
In this paper we propose a strategy to quantify the covariate shift in a temporal dataset. This strategy consists in classifying the transactions of each day against every other days: If the classification is efficient then the days are different and there is a covariate shift between them. On the other hand, if the classification is not efficient, the days are similar. This strategy allows us to build a distance matrix characterizing the covariate shift between days. 

Afterwards, we use an agglomerative clustering algorithm on the distance matrix between days. We show that in the case of the belgian face-to-face credit card transactions the dataset shift matches almost exactly the calendar events. We identify 4 types of days in the credit card transactions dataset: 'working days', 'saturdays', 'sundays' and 'school holiday'. 

Using a Random Forest classifier, we show that integrating the information of the type of the day previously identified increases the Precision-Recall AUC by a small percentage (2.5\%).

\bibliographystyle{splncs}


\begin{figure*}
\centering
\includegraphics[width=0.76\textwidth]{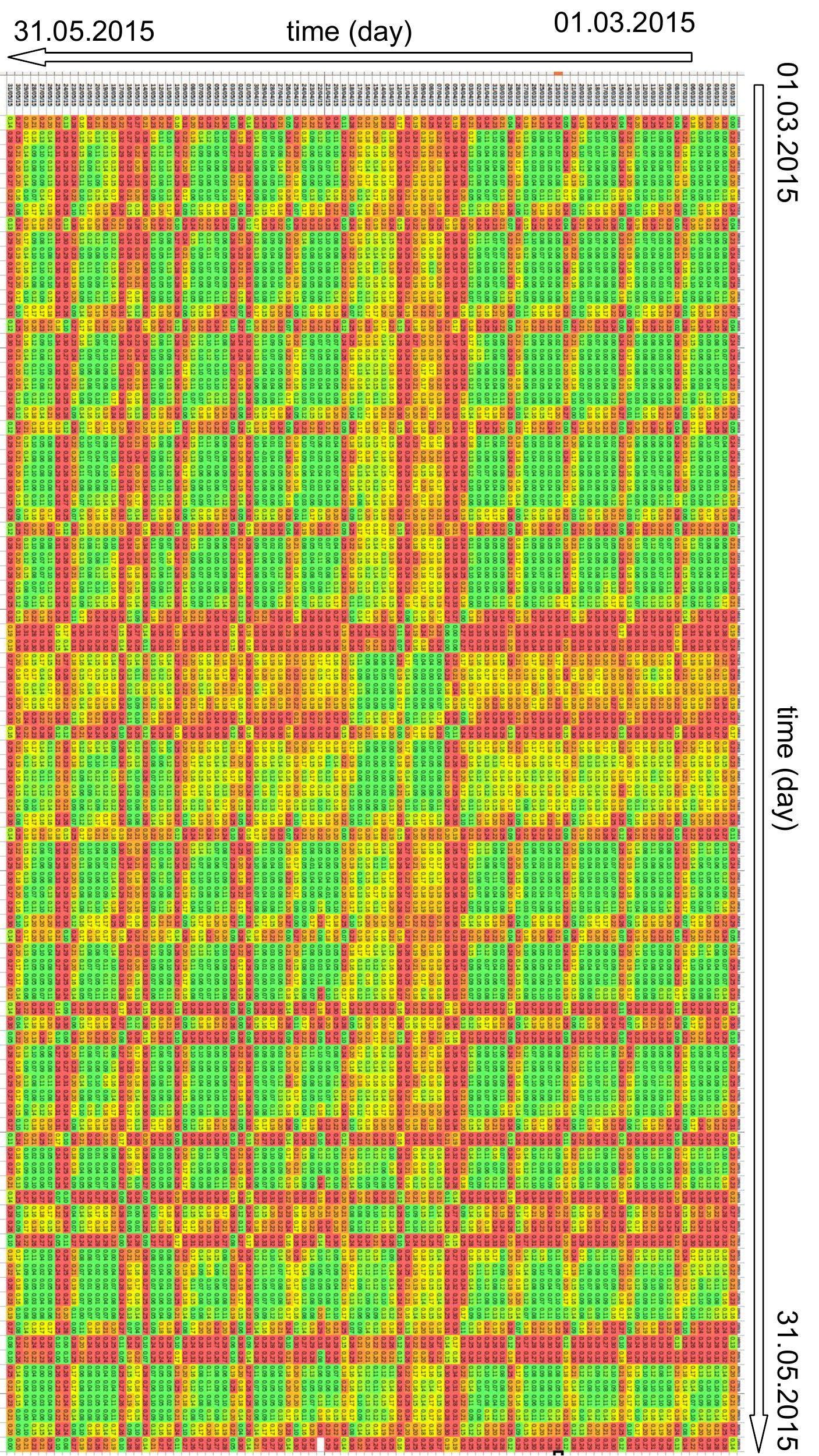}
\caption{Appendix: distance matrix (dataset shift) for the face-to-face transactions. \textit{(centile based coloration: green $\Leftrightarrow$ similar days (MCC$\approx$0), red $\Leftrightarrow$ dataset shift (MCC$\approx$1)).}
We observe a strong periodicity in the patterns detected with our dataset shift quantification strategy}
\label{dist_mat}
\end{figure*}

\end{document}